\documentclass[runningheads]{llncs}
\pdfoutput=1
\usepackage[backref=page]{hyperref}
\usepackage{amsmath}
\usepackage{url}
\usepackage{bm}
\usepackage{graphicx}
\usepackage{subfigure}
\usepackage{comment}
\usepackage{amsmath,amssymb} 
\usepackage{color}
\usepackage{multirow}
\usepackage{booktabs}
\usepackage{array}
\usepackage{lineno,hyperref}
\usepackage[misc]{ifsym}
\usepackage[linesnumbered,boxed,ruled,commentsnumbered]{algorithm2e}

\begin{document}
\pagestyle{headings}
\mainmatter
\def\ECCVSubNumber{6114}  

\title{Improving the Transferability of Adversarial
	Examples with New Iteration Framework and Input Dropout}

\titlerunning{Improving the Transferability with NIF and ID}
%

\author{Pengfei Xie\inst{1} \and
 Linyuan Wang\inst{1} \and
Ruoxi Qin\inst{1} \and
Kai Qiao\inst{1} \and
 Shuhao Shi\inst{1} \and
 Guoen Hu\inst{1} \and
Bin Yan\inst{1}
 \thanks{Corresponding author}}
\index{Shen, Heng Tao}
\authorrunning{P.Xie, L.Wang et al.}
%

\institute{ Henan Key Laboratory of Imaging and Intelligent Processing, PLA Strategy Support Force Information Engineering University, Zhengzhou, China 
\\
\email{ybspace@hotmail.com}}
\maketitle
\begin{abstract}
		Deep neural networks(DNNs) is vulnerable to be attacked by adversarial examples. Black-box attack is the most threatening attack. At present, black-box attack methods mainly adopt gradient-based iterative attack methods, which usually limit the relationship between the iteration step size, the number of iterations, and the maximum perturbation. In this paper, we propose a new gradient iteration framework, which redefines the relationship between the above three. Under this framework, we easily improve the attack success rate of DI-TI-MI-FGSM. In addition, we propose a gradient iterative attack method based on input dropout, which can be well combined with our framework. We further propose a multi dropout rate version of this method. Experimental results show that our best method can achieve attack success rate of 96.2\% for defense model on average, which is higher than the state-of-the-art gradient-based attacks.

\end{abstract}

\section{Introduction}
\label{Introduce}
	In recent years, deep neural networks have begun to shine. It has achieved good results in tasks such as image classification \cite{he_deep_2016,szegedy_rethinking_2016} object detector \cite{redmon_yolov3_2018}, speech recognition \cite{hinton_deep_2012}, and natural language processing \cite{sutskever_sequence_2014}. However, studies have shown that adversarial example, with imperceptible and carefully designed noise added to the original image, can make DNNs wrong. This has attracted great attention from the academic community. In order to enable the safe deployment of applications such as autonomous driving and face recognition, it is necessary to study attack algorithms and defense methods about adversarial examples, and the two promote and develop each other.\\
At present, adversarial attacks are mainly divided into digital domain attacks and physical domain attacks. 
\begin{figure}
	\centering
	\includegraphics[height=5cm]{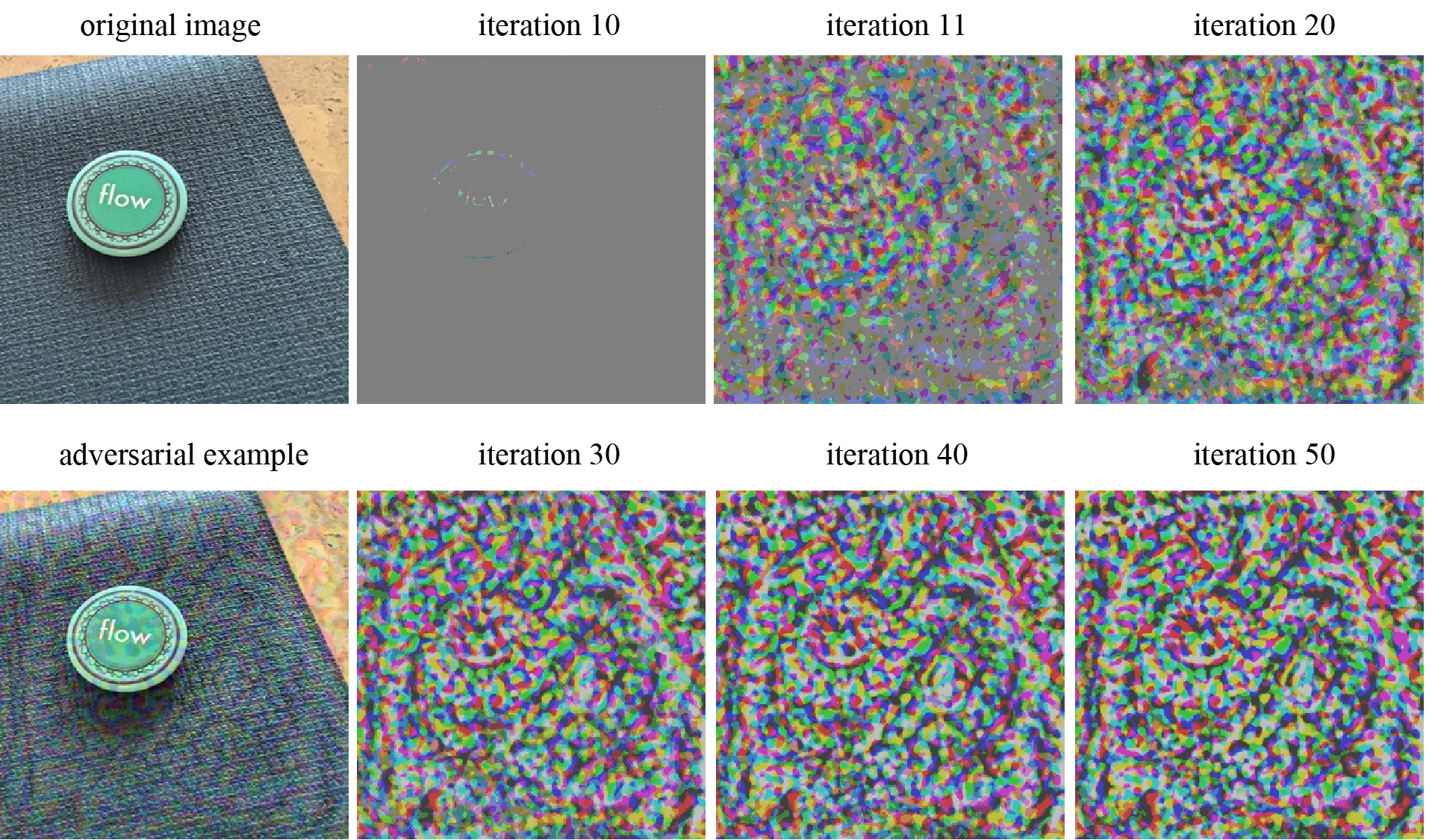}
	\caption{We show the spillover disturbance under clip with the step size as 1.6 and the number of iterations as 50. Our adversarial examples are generated by DI-TI-MI-FGSM.}
	\label{icml-clip}
\end{figure}
Digital domain attacks are the basis of physical domain attacks. Robust digital domain attack algorithms can be effectively used to physical domain attacks through 3D printing \cite{athalye_synthesizing_2018} and light projection \cite{nguyen_adversarial_2020}. The attack methods for generating adversarial examples can be divided into methods based on optimization \cite{moosavi-dezfooli_deepfool_2016,carlini_towards_2017},gradient iteration \cite{goodfellow_explaining_2014,kurakin_adversarial_2016} and generative model \cite{baluja_adversarial_2017,xiao_generating_2018}. Among them, gradient-based attack methods are the most popular because of their fast generation speed, low algorithm complexity and good transferability, i.e., the adversarial examples generated by the gradient information of the substitute model also have a high attack success rate on other models. This type of attack poses a huge security threat on deep neural network, because the attacker does not need to know any information about the attacked model but make it wrong. The gradient-based attack method are developed from one-step gradient approaches Fast Gradient Sign Method (FGSM) , and then developed into multi-step iterative approaches Iterative Fast Gradient Sign Method (I-FGSM). In the case of white-box, the attack success rate of I-FGSM is usually better than that of FGSM, because it finds the direction that maximizes the loss function more precisely through multi-step iteration. But under the black-box setting, I-FGSM is usually weaker than FGSM, because it easily falls into overfitting in the substitute model, which makes it difficult to transfer the attack performance to other models. In order to make this kind of method have good performance in both white-box setting and black-box setting, some work \cite{dong_boosting_2018,xie_improving_2019,dong_evading_2019} begin to study how to improve transferability of adversarial examples.\\
In previous work, in order to avoid gradient overflow, step size, iteration times and maximum disturbance often satisfy the certain relationship. In this paper, we further discuss the relationship among the three and propose a new gradient iteration framework. We improve the transferability of adversarial examples by a large constant step size and increasing the number of iterations. We use this method to easily improve the attack performance of Diverse Input, Translation Invariant, and Momentum Iterative Fast Gradient Sign Method(DI-TI-MI-FGSM).\\
In order to further improve the transferability of adversarial examples, we propose an Input Dropout Iterative Fast Gradient Sign Method (IDI-FGSM). Different from previous work \cite{li_learning_2020}, we work at the data level rather than at the network level. We find that the classification accuracy and loss of the original image basically remain unchanged with the dropout rate transformation, so we believe that dropout is invariant, which is similar to the previous work \cite{dong_evading_2019,xie_improving_2019,lin_nesterov_2019}. However, dropout will lose part of the information of the adversarial examples, affecting its attack performance. Based on this, we introduce dropout into the gradient iteration attack method. Our method is easy to implement and can be effectively combined with other methods. In addition, we propose an Input Dropout Ensemble Iterative Fast Gradient Sign Method (IDEI-FGSM), i.e., a multiple dropout rate version of IDI-FGSM to further improve the transferability of adversarial examples. The experimental results show that our best method achieves an average attack success rate of 96.2\% on the defense model.
	In summary, we make the following contributions:
	\begin{itemize}
	\item[$\bullet$] We study the relationship between the step size and the number of iterations in the gradient iteration attack method, and propose a new gradient iteration framework. Unlike previous frameworks, this framework allows gradient overflow.   We use this framework to improve the attack performance of DI-TI-MI-FGSM easily. Our framework can be well combined with our proposed input dropout method. We explain why constant step size and increased iterations can improve attack performance.
	\item[$\bullet$] We propose an Input Dropout Iterative Fast Gradient Sign Method (IDI-FGSM), which can be well combined with other methods based on gradient iteration, and achieve an average attack success rate of 95.3\% on the defense model.
	\item[$\bullet$] We further propose a gradient iterative attack method with multiple dropout rate versions, i.e., Input Dropout Ensemble Iterative Fast Gradient Sign Method (IDEI-FGSM), which can achieve 96.2\% success rate in the defense model.
	\end{itemize}

\section{Related work}
	Szegedy et at. \cite{szegedy_intriguing_2013} first propose adversarial examples. They use the L-BFGS box constraint algorithm to maximize the loss function between the original sample and ground-truth label, and constraint the disturbance size to obtain the adversarial examples, which can make the DNNs wrong. Subsequently, a large number of digital domain adversarial attack algorithms have been proposed, including optimization methods, gradient iteration methods, and generative model methods. Accordingly, some defense methods are proposed accordingly \cite{tramer_ensemble_2017,xie_feature_2019}. At present, the most effective defense method is adversarial training \cite{ganin_domain-adversarial_2016}. However, some studies have shown that adversarial examples not only exist in the digital domain, but also can be applied to the real world. Advhat \cite{komkov_advhat_2021} successfully deceives the face recognition model, and Adv-Thirt \cite{xu_adversarial_2020} successfully deceives the object detection model. More and more studies have shown that adversarial examples exist not only in computer vision, but also in speech recognition \cite{carlini_audio_2018}, reinforcement learning \cite{behzadan_vulnerability_2017}, and text model \cite{jin_is_2020}. It should be noted that our work is based on image classification tasks.
\subsection{White-box attack}
White-box attack refers to that an attacker can obtain all information of the model including network structure and parameters. These attack methods include optimization-based attacks, such as DeepFool \cite{moosavi-dezfooli_deepfool_2016}, CW attacks \cite{carlini_towards_2017}. However, the computation time of these methods is too long and it is easy to lead to overfitting, which makes it difficult to transfer attack performance to other models. 
In addition, there are some attack methods based on generative model, such as ATN \cite{baluja_adversarial_2017}, AdvGAN \cite{xiao_generating_2018}, which learn the distribution of adversarial examples by training a generator. However, this method require a lot of time and data to train a generator.
The method based on gradient iteration has the characteristics of good attack performance, easy implementation of code and fast operation speed. This kind of method is the mainstream algorithm against attack.

\subsection{Black-box attack}
Black-box attack refers to the attacker cannot obtain the network structure, parameters and other information of the model. It can be subdivided into semi-black-box attacks and completely black-box attacks. In a semi-black box attack scenario, an attacker can estimate its gradient by querying the returned value by the model, such as the Zoo attack \cite{chen_zoo_2017}. In a black box scenario, attackers attack substitute models to generate adversarial examples, and then use its transferability to attack other models. Dong et at. \cite{dong_boosting_2018} propose an iterative algorithm based on momentum to enhance adversarial attacks. In addition, these works improve the transferability of adversarial examples by various transformations \cite{dong_evading_2019,xie_improving_2019,lin_nesterov_2019}. Wu et at. \cite{wu_understanding_2018} smooth the gradient of local area to increase transferability. Gao et at. \cite{ssm} replace the traditional sign method with their staircase sign method to effectively utilize the gradient of the substitute model, thereby improving the transferability of adversarial examples. In this paper, our goal is to improve the transferability of adversarial examples in black box backgrounds.

\section{Methodology}
	In this section, we provide a detailed description of our algorithm. Let $x$ denote the clean example, an ${y^{true}}$ denote the corresponding label. We use $F$ to represent the DNNs, which can correctly map $x$ to $y$, namely $F(x) = {y^{true}}$. Then we create adversarial examples ${x^{adv}}$ that appears to be similar to the clean example, which can make the DNNs wrong, i.e., $F(x) \ne {y^{true}}$, and ${\left\| {x - {x^{adv}}} \right\|_p} \le \varepsilon $, where $p$ usually as $0,1,2,\infty $ and $\varepsilon $ as the maximum value of the adversarial perturbation. To generate adversarial examples, we maximize the cross entropy loss function $J({x^{adv}},{y^{true}})$, which will greatly reduce the confidence of adversarial examples on the true label, and cause classification errors. The optimization objective function are as follows:
\begin{equation}
\ \mathop {\arg \max }\limits_{{x^{adv}}} J({x^{adv}},{y^{true}}),  {\rm{s}}.{\rm{t}}.{\left\| {{x^{adv}} - x} \right\|_\infty } \le \varepsilon .\
\end{equation}
	Following previous work \cite{dong_evading_2019,xie_improving_2019}, we use ${L_\infty }$ norm constrain the size of the disturbance. Due to the lack of internal information of the attacked model in black box settings, we use transferability to attack. How to improve the transfer attack ability between models is a challenging work. Below we will also introduce the classic algorithm of improving the ability of black box transfer attack.
\subsection{Gradient-Based Attack Methods}
In this section, we briefly introduce some excellent works based on gradient sign method.\\
\textbf{Fast Gradient Sign Method (FGSM):} Goodfellow et at. \cite{goodfellow_explaining_2014} believe that the vulnerability of neural network stems from its linear nature, and proposed a gradient-based one-step attack method for the first time. The formula of updating adversarial examples is as follows:
\begin{equation}
\ {x^{adv}} = {x^{real}} + \varepsilon  \cdot sign({\nabla _x}J({x^{real}},{y^{true}})),\
\end{equation}
\textbf{Iterative Fast Gradient Sign Method (I-FGSM):} Kurakin et at. \cite{kurakin_adversarial_2016} propose a multi-step iterative version of FGSM, which reaches the maximum loss point of the model with a smaller step.  The updated formula is as follows:
\begin{equation}
\ X_0^{adv} = {X^{real}},\ 
X_{t + 1}^{adv} = Clip_X^\varepsilon \{ X_t^{adv} + \alpha  \cdot sign({\nabla _{X_t^{adv}}}J(X_t^{adv},{y^{true}}))\} ,  
\end{equation} 
where $Clip$ denotes the clipping operation of adversarial examples for each iteration to limit adversarial  perturbations into the ${L_\infty }$ norm $\varepsilon $.\\
\textbf{Momentum Iterative Fast Gradient Sign Method (MI-FGSM):} Dong et at. \cite{dong_boosting_2018} introduce momentum into the gradient iteration process to stabilize the gradient update direction and avoid falling into extreme points. The formula is as follows:
\begin{equation}
\ {g_{t + 1}} = \mu  \cdot {g_t} + \frac{{{\nabla _x}J(x_t^{adv},{y^{true}})}}{{{{\left\| {{\nabla _x}J(x_t^{adv},{y^{true}})} \right\|}_1}}},\  
X_{t + 1}^{adv} = Clip_X^\varepsilon \{ X_t^{adv} + \alpha  \cdot sign({g_{t + 1}})\} ,
\end{equation}
\textbf{Diverse Input Iterative Fast Gradient Sign Method (D${{\rm{I}}^2}$-FGSM):} Xie et at. \cite{xie_improving_2019} propose input transformation diversity to improve the mobility of adversarial examples. This method can be written as:
\begin{equation}
\ X_{t + 1}^{adv} = Clip_X^\varepsilon \{ X_t^{adv} + \alpha  \cdot sign({\nabla _{X_t^{adv}}}J(T(X_t^{adv},p),{y^{true}}))\} \ 
\end{equation}
where $T$ denote input transformation with $p$.  \\
	\textbf{Translation-Invariant Attack Method (TI-FGSM):} Dong et at. \cite{dong_evading_2019} generate adversarial examples insensitive to the white-box discriminative region by translation invariance, and use predefined convolution kernels to replace translation operations to improve efficiency.  This method has good attack performance on defense model.\\
\textbf{Scale-Invariant Nesterov Iterative Fast Gradient Sign Method (SI-NI-FGSM):} Lin et at. \cite{lin_nesterov_2019} adopt Nesterov accelerated gradient into the iterative attacks to effectively escape from global maximum. In addition, they also use the scale invariance of DNNs to improve the transferability of adversarial examples.\\
\textbf{Patch-wise Iterative Fast Gradient Sign Method (PI-FGSM):} Gao et at. \cite{gao_patch-wise_2020} adopt amplification operation on the step size and use the pre-trained convolution kernel to project the overflowed gradient information to the surrounding area to improve transferability. \\
\textbf{Resized-Diverse-Inputs Diversity-Ensemble Iterative Fast Gradient Sign Method (RDI-DEI-FGSM):} Zou et at. \cite{zou_improving_2020} find that there are many vertical and horizontal stripes in the gradients of diverse inputs, which can be used to alleviate the loss of gradient information caused by TI-FGSM. They propose that resized-diverse-inputs method (RD${{\rm{I}}^2}$-FGSM) can combine with TI-FGSM to play a better attack performance. In addition, they propose diversity-ensemble method (DEM), that is, i.e., the multi-scale version of RD${{\rm{I}}^2}$-FGSM to further improve the transferability of adversarial examples.
\begin{figure}
	\centering
	\includegraphics[width=8cm]{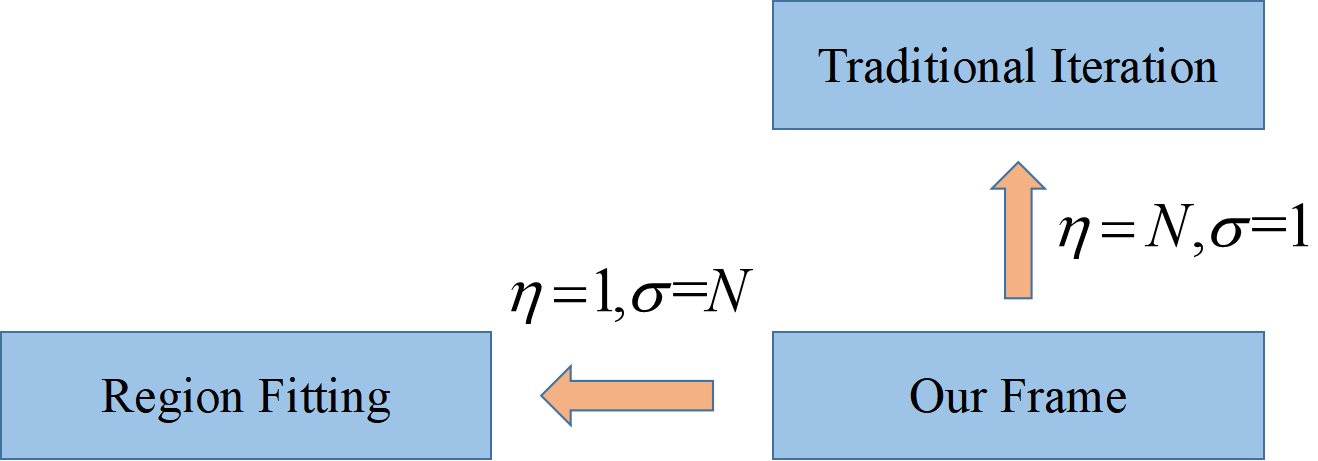}
	\caption{We show the relationship between our framework and previous iteration methods.}
	\label{icml-frame}
\end{figure}

\subsection{New iteration framework}
As we know, the current attack methods based on gradient iteration usually limit the infinite norm disturbance $\varepsilon $, iteration number $T$ and iteration step size $\alpha $ to the following formulas for preventing gradient overflow.
\begin{equation}
\alpha  = \frac{\varepsilon }{T}\ 
\end{equation}
Zou et at. \cite{zou_improving_2020} propose the regional Region Fitting, which expand the iteration step size $\alpha $ to $\varepsilon $, but its iteration times are still limited. Gao et at. \cite{gao_patch-wise_2020} propose to project the overflow gradient information to the regional patch to improve transferability. Our method is different from the above method, and it is very direct. Region Fitting can be seen as a case in our method. We directly select a large step size to prevent over-fitting, and then extend the number of iterations to improve attack performance. We allow gradient overflow, because in a certain iteration round, the benefit of extending the iteration round is greater than the loss of attack performance caused by clip. 
We propose a new gradient iteration framework, which can be expressed as:
\begin{equation}	
\ \left \{	
\begin{array}{ll}	
{\alpha  = \frac{\varepsilon }{\eta}, }                  & { \eta \ge 1 }\\
{T = \sigma  \cdot \eta, }     &   { \sigma \ge 1 }  \quad {and} \quad {T \in {N^ + }}     
\end{array}	
\right.	\ 
\end{equation} \\
where $\eta $ is the step size expansion factor, which determine the step size, $\sigma $ is the iterative times factor. As shown in Figure~\ref{icml-frame}, We show the relationship between our framework and the traditional gradient iteration method and Region Fitting.

The gradient descent leads to the increase of loss, while the clip limits the disturbance to reduce loss, and the two are in a game state. As shown in Figure~\ref{icml-clip}, we show the perturbations that overflow under different iterations. It can be seen that the disturbance does not overflow in the 10th round, and the disturbance begins to overflow obviously in the 11th round. However, the disturbance of clip is relatively small and the defense performance is weak. At this time, the attack performance still has room for improvement. As the number of iterations increases, the perturbations of clips increase gradually and begin to show a fixed shape, which may be a balanced state for the gradient descent and clip.
\begin{figure}
	\centering
	\includegraphics[width=\linewidth]{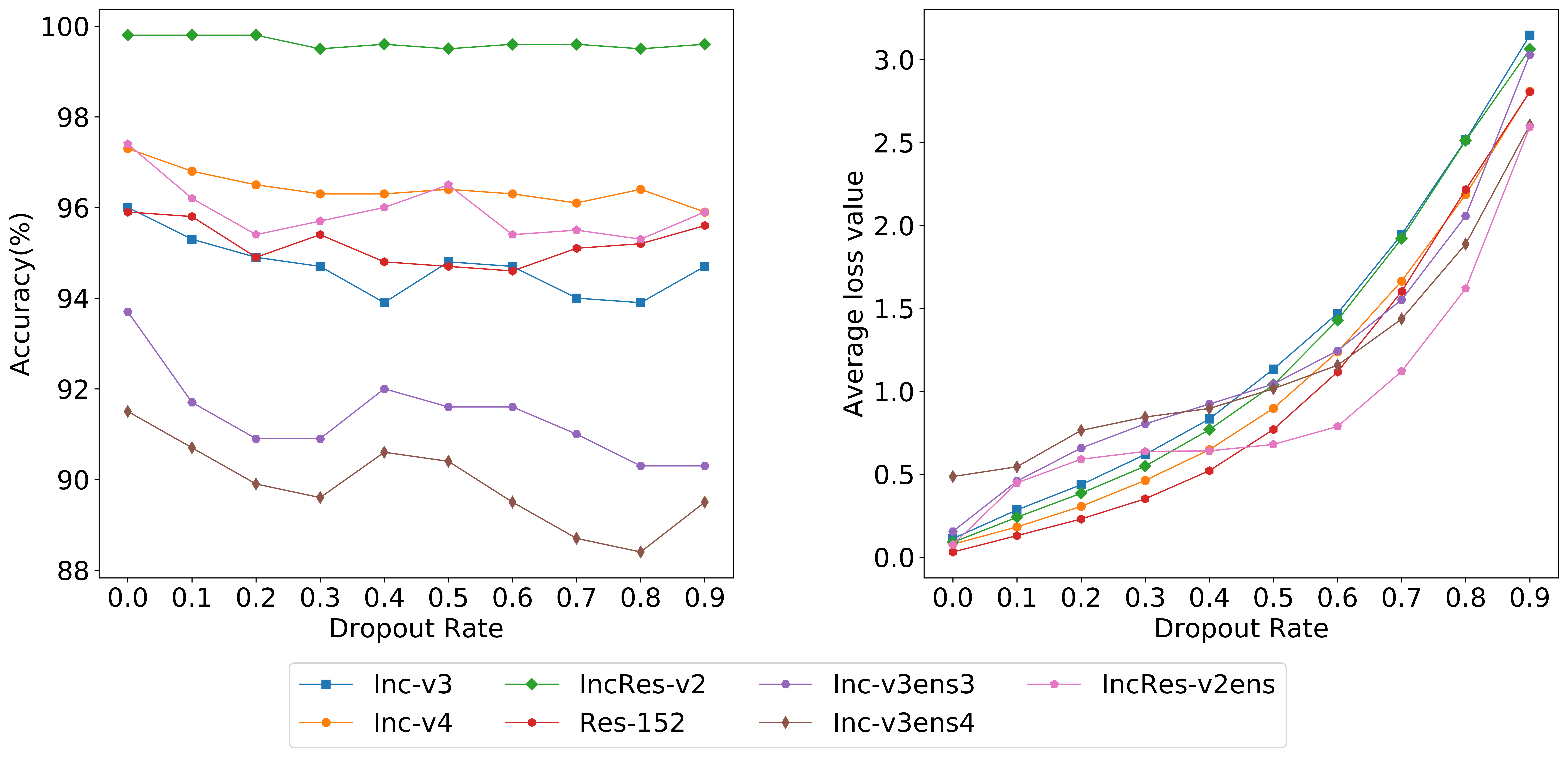}
	\caption{The curves of accuracy and average loss value for Inc-v3, Inc-v4, IncRes-v2, Res-152, Inc-v${{\rm{3}}_{ens3}}$, Inc-v${{\rm{3}}_{ens4}}$ and IncRes-v${2_{{\rm{ens}}}}$ given the dropout images at each dropout rate. The results are averaged over 1000 images.}
	\label{icml-dropout_nochange}
\end{figure}

\subsection{Dropout}
In this section, we will describe our dropout-based method in detail.
\subsubsection{Motivation}
Dropout \cite{srivastava_dropout_2014} is a commonly used regularization method, which can effectively prevent network over-fitting and enhance the generalization performance of the network. Inspired by this, we introduce dropout into the generation process of adversarial examples, which can prevent the generated adversarial examples from overfitting on the substitute model to enhance the transferability of adversarial examples. \\
From the perspective of data enhancement, dropout reduces the attack performance of adversarial examples and turns part of them into correctly classified examples, which makes the data more diverse. These examples will be used in the gradient iterative process to improve the transferability of adversarial examples.   
From the perspective of attack and defense, dropout can be regarded as a defense mechanism, which can make the adversarial examples lose part of the information, thereby reducing the attack effect of the adversarial examples. In the attack stage, the adversarial examples will adaptively overcome this defense mechanism during gradient descent, and become a more robust adversarial examples. \\
Wang et at. \cite{wang_defensive_2018} use dropout in the test, which can effectively reduce the success rate of adversarial examples. Li et at. \cite{li_learning_2020} makes the dropout layer of the substitute model more diversified to increase the transferability of adversarial examples. Above work are carried out around the dropout layer in the network structure, which must require the existence of dropout in the network, and our work is carried out around the data level, and there is no requirement for the network structure. 
Similar to translation-invariant \cite{dong_evading_2019} and scale-invariant \cite{lin_nesterov_2019}, we verify that dropout also has invariance. As shown in Figure~\ref{icml-dropout_nochange}, when the dropout rate is between 0-0.5, the loss value and classification accuracy of the original image remain basically unchanged.

\IncMargin{1em} 
\begin{algorithm}[t]
	\DontPrintSemicolon
	\SetAlgoNoLine 
	\SetKwInOut{Input}{\textbf{Input}}\SetKwInOut{Output}{\textbf{Output}}
	\Input{ A clean example $x$ normalized to $[ - 1,1]$; ground-truth label ${y^{true}}$ ; ${L_\infty }$ norm constraint $\varepsilon $; step size $\alpha $ ; iteration times $T$; a substitute model $f$ with loss function $J$; dropout rate $\gamma $.
	}
	\Output{The adversarial example ${x^{adv}}$.
	}
	Initialize $x_0^{adv} = x$;\quad\\	
	\For{$t \leftarrow 0$ \KwTo $T-1$}
	{
		$x_t^{adv} = D{\rm{rop}}(x_t^{adv},(1 - \gamma )) \cdot (1 - \gamma ))$;\quad\\
		Get the gradients by ${\nabla _{x_t^{adv}}}J(x_t^{adv},{y^{true}})$; \quad\\
		Update adversarial example ${\nabla _{x_t^{adv}}}J(x_t^{adv},{y^{true}})$
		Clip adversarial example $x_{t + 1}^{adv} = Clip(x_{t + 1}^{adv}, - 1,1)$
	}
	Return  ${x^{adv}} = x_t^{adv}$;
	\caption{IDI-FGSM\label{IDI-FGSM}}
\end{algorithm}
\DecMargin{1em}

\subsubsection{IDI-FGSM}
We propose IDI-FGSM, which is summarized in Algorithm \ref{IDI-FGSM} . Our method is intuitive and easy to implement, and can be easily integrated into other gradient-based iterative methods. Specifically, before we input the adversarial examples into the neural network, we will dropout the adversarial examples, that is, randomly set 0 for some pixels of adversarial examples, where dropout rate takes [0,1). Our method can alleviate the over-fitting phenomenon of adversarial examples in the substitute model, so as to improve the transferability of adversarial examples. Our method is iterative by the following methods:
\begin{equation}
\;X_{t + 1}^{adv} = Clip_X^\varepsilon \{ X_t^{adv} + \alpha  \cdot sign({\nabla _{X_t^{adv}}}J((D{\rm{rop}}(X_t^{adv},(1 - \gamma )) \cdot (1 - \gamma )),{y^{true}}))\} {\rm{ }}
\end{equation}
where $\gamma $ denotes dropout rate.

\subsubsection{IDEI-FGSM}
We further propose a gradient iterative attack method with multiple dropout rate versions, i.e., IDEI-FGSM, which is summarized in Algorithm \ref{IDEI-FGSM} . Specifically, we use different dropout rate in the dropout stage, then integrate the logit obtained by different dropout rate to the cross-entropy loss function. Compared with the single version of dropout rate, this method can obtain more abundant and good examples when the gradient descent, so it has stronger transferability. Our formula of logit is as follows:
\begin{equation} 
\;l(x) = \sum\nolimits_1^K {{\omega _k}}  \cdot l(Drop(x_t^{adv},(1 - {\gamma _k})) \cdot (1 - {\gamma _k}))\  \label{eqa:9}
\end{equation}
where ${\sum\nolimits_1^K \omega  _k} = 1$, ${\omega _k} \ge 0$, ${\gamma _k} \in [0,1)$.
Following the former work, we use the softmax cross-entropy loss as our loss function:
\begin{equation}
\;J(x_t^{adv},{y^{true}}) =  - {1_{{y^{true}}}} \cdot \log ({\rm{softmax}}(l(x_t^{adv})))     \label{eqa:10}
\end{equation}	

\IncMargin{1em} 
\begin{algorithm}[t]
	\DontPrintSemicolon
	\SetAlgoNoLine 
	\SetKwInOut{Input}{\textbf{Input}}\SetKwInOut{Output}{\textbf{Output}}
	\Input{ A Clean example $x$ normalized to $[ - 1,1]$; ground-truth label ${y^{true}}$ ; ${L_\infty }$ norm constraint $\varepsilon $; step size $\alpha $ ; iteration times $T$; a substitute model $f$ with loss function $J$; weights ${\omega _1},{\omega _1} \ldots ,{\omega _k}$; the logits of $K$  dropout rate 
		$l(D{\rm{rop}}(X_t^{adv},(1 - {\gamma _1})) \cdot (1 - {\gamma _1}))$,$l(D{\rm{rop}}(X_t^{adv},(1 - {\gamma _2})) \cdot (1 - {\gamma _2}))$,$ \ldots $,$l(D{\rm{rop}}(X_t^{adv},(1 - {\gamma _K})) \cdot (1 - {\gamma _K}))$.
	}
	\Output{The adversarial example ${x^{adv}}$.
	}
	Initialize $x_0^{adv} = x$;\quad\\	
	\For{$t \leftarrow 0$ \KwTo $T-1$}
	{
		Input $x_t^{adv}$ and Get logits $l(x_t^{adv})$  following \ref{eqa:9};\quad\\
		Get cross-entropy loss $J(x_t^{adv},{y^{true}})$ by ${\nabla _{x_t^{adv}}}J(x_t^{adv},{y^{true}})$   by \ref{eqa:10}; \quad\\
		Obtain the gradients by ${\nabla _{x_t^{adv}}}J(x_t^{adv},{y^{true}})$
		Update adversarial example ${\nabla _{x_t^{adv}}}J(x_t^{adv},{y^{true}})$
	    Clip adversarial example $x_{t + 1}^{adv} = Clip(x_{t + 1}^{adv}, - 1,1)$
	}
	Return  ${x^{adv}} = x_t^{adv}$;
	\caption{IDEI-FGSM\label{IDEI-FGSM}}
\end{algorithm}
\DecMargin{1em}

\begin{table}[t]
	\centering
	\caption{Abbreviations used in the paper.}
	\resizebox{\textwidth}{!}{
			\begin{tabular}{c|c|}
			\toprule
			Abbreviation&Explanation	\\
			\midrule
			DI-TI-MI-FGSM&	The combination of D${{\rm{I}}^2}$-FGSM,TI-FGSM,and MI-FGSM \\
			
			DI-TI-PI-FGSM&	The combination of D${{\rm{I}}^2}$-FGSM,TI-FGSM,and PI-FGSM  \\
			
			DI-TI-ID-PI-FGSM&	The combination of D${{\rm{I}}^2}$-FGSM,TI-FGSM,PI-FGSM,and IDI-FGSM  \\
			
			DI-TI-ID-MI-FGSM&	The combination of D${{\rm{I}}^2}$-FGSM,TI-FGSM,PI-FGSM and IDI-FGSM  \\
			
			DI-TI-IDE-MI-FGSM&	The combination of D${{\rm{I}}^2}$-FGSM,TI-FGSM,MI-FGSM and IDEI-FGSM \\
			\bottomrule
			\end{tabular}}
	\label{sample-table3}%
\end{table}%

\section{Experiments }
In this section, we conduct experiments on the ImageNet dataset to verify the effectiveness of our method. We specify the experimental setup parameters in Sec.~\ref{sec:1} . We test our iteration framework in Sec.~\ref{sec:2} . We show the success rate of black box attacks under different dropout rates in Sec.~\ref{sec:3} . We report the results of attacking a single model in Sec.~\ref{sec:4} and an ensemble of	models in Sec.~\ref{sec:5} .

\subsection{Experimental Settings}\label{sec:1}
\textbf{Dataset.}  Following the previous work \cite{dong_boosting_2018,dong_evading_2019}, we conduct experiments on the Imagenet dataset, including 1000 images that are resized to 299×299×3, which are used in the competition of NIPS2017. \\
\textbf{Networks.} We selected ten models for experiments, including four normal trained models, i.e., Inception-v3(Inc-v3) \cite{szegedy_rethinking_2016}, Inception-v4 (Inc-v4), Inception-Resnet-v2(IncRes-v2), \cite{szegedy_inception-v4_2017} Resnet-v2-152 (Res-152) \cite{he_deep_2016}, and three ensemble adversarially trained models \cite{tramer_ensemble_2017}, i.e., ens3-adv-Inception-v3 (Inc-v${{\rm{3}}_{ens3}}$), ens4-adv-Inception-v3 (Inc-v${{\rm{3}}_{ens4}}$) and ens-adv-Inception-ResNet-v2 (IncRes-v${2_{{\rm{ens}}}}$), and three more robust denoise models \cite{xie_feature_2019}, ResNet152 Baseline (Res15${2_B}$), ResNet152 Denoise (Res15${2_D}$), ResNeXt101 DenoiseAll (ResNeX${{\rm{t}}_{DA}}$).\\
\textbf{Implementation details.} In our experiment, we compare the advanced attack methods based on gradient iteration, including I-FGSM, MI-FGSM, PI-FGSM, D${{\rm{I}}^2}$-FGSM, TI-FGSM, and their combinations, namely DI-TI-MI-FGSM and DMPI-FGSM. In all experiments, our maximum perturbation $\varepsilon $ is set 16. For the previous work sucn as I-FGSM, D${{\rm{I}}^2}$-FGSM, PI-FGSM, DI-TI-MI-FGSM and DTPI-MI-FGSM, we set the number of gradient iterations to 10 and the step size to 1.6. 
For MI-FGSM, we set the decay factor $\mu  = 1.0$. For TI-BIM, we set the kernel size $k = 15$.  and for D${{\rm{I}}^2}$-FGSM, we set the transformation probability $p = 0.7$. For PI-FGSM, we set amplifification factor $\beta  = 10$. 
For IDI-FGSM, we set the dropout rate to 0.1. For IDEI-FGSM, we set the dropout rate to 0.0, 0.1, 0.2, 0.3, 0.4, and the weight factors are equal. When attacking the ensemble of models, we set the equal weight.

\subsection{Extend the number of iterations}\label{sec:2}
In this section, we focus on the relationship between step size and iteration round. We attack the Inc-v3 model with DI-TI-MI-FGSM algorithm to generate adversarial examples, and then test them on both defenseless model: Inc-v4, IncRes-v2, and Res-152, and defense models: Inc-v${{\rm{3}}_{ens3}}$, Inc-v${{\rm{3}}_{ens4}}$, and IncRes-v${2_{{\rm{ens}}}}$. We set the iteration round to 50, and test the attack success rate under different iteration times, where the step size is set to 1.6, 3.2, 4.8, 6.4, 8.0, 9.6, 11.2, 12.8, 14.4, 16.0 respectively. Specifically, we further study DI-TI-MI-FGSM and DI-TI-ID-MI-FGSM algorithm with the case of step size as 1.6 and 16, and extend the iteration rounds to 200 times. \\
 \begin{figure}
	\centering
	\includegraphics[width=\linewidth]{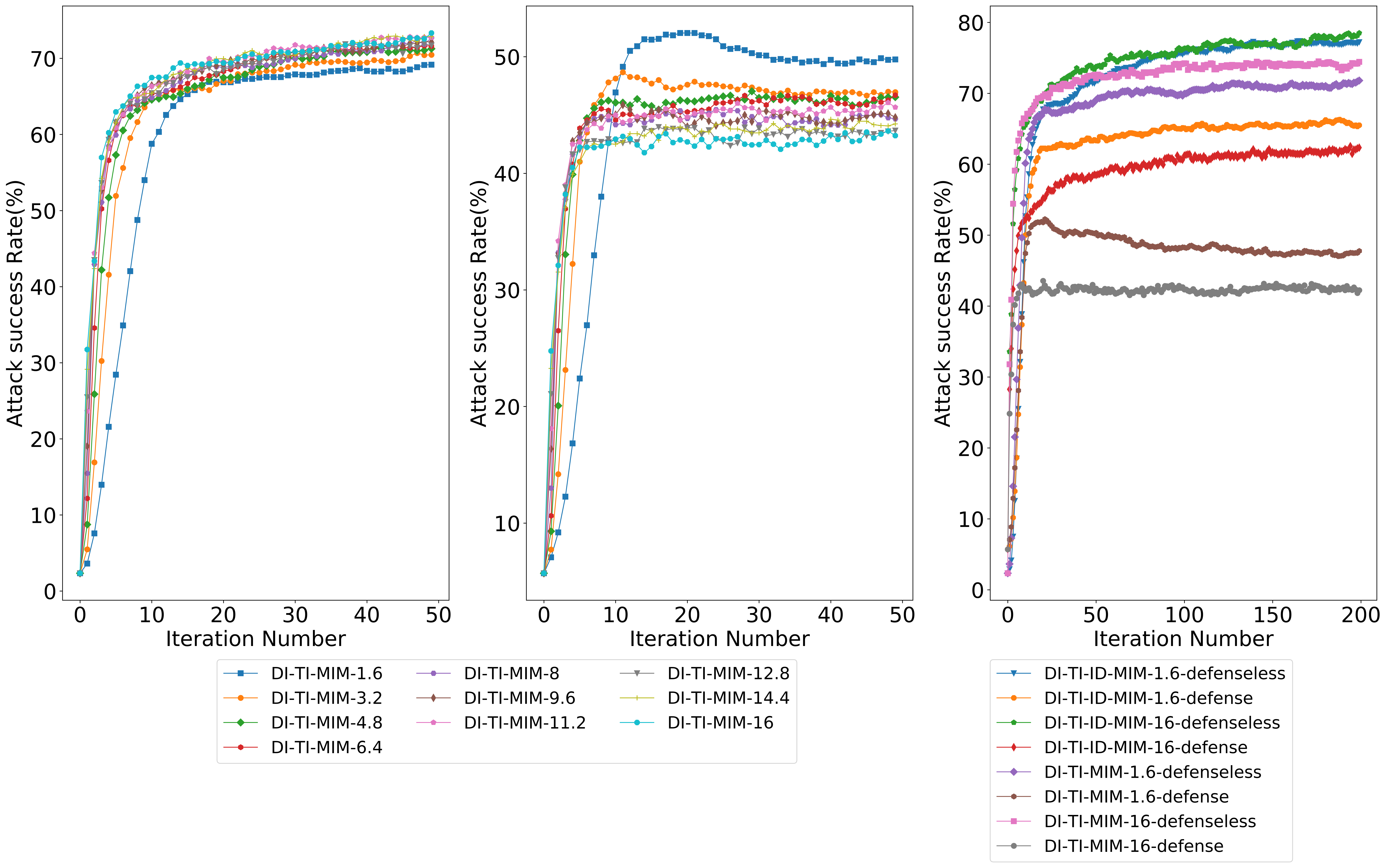}
	\caption{The curves of average attack success rate $(\% )$ for defenseless model: Inc-v4, IncRes-v2, Res-152 and defense model: Inc-v${{\rm{3}}_{ens3}}$, Inc-v${{\rm{3}}_{ens4}}$, IncRes-v${2_{{\rm{ens}}}}$. The adversarial examples are crafted for Inc-v3 model by DI-TI-MI-FGSM and DI-TI-ID-MI-FGSM. Left Column: The average attack success rate of three defenseless model with 10 different step sizes. Middle Column: The average attack success rate of three defense model with 10 different step sizes. Right Column: The average attack success rate of three defenseless models and three defense models with step sizes as 1.6 and 16.}
	\label{icml-iteration}
\end{figure}
As shown in the as in Figure~\ref{icml-iteration}, we find that when the number of iterations is extended, the attack success rate can be significantly improved. Compared with 10 iterations, the attack success rate of 50 iterations with different step size increases by an average of 6.9\% on the defenseless model. When the step size is larger, the convergence of loss is faster, and it can achieve a higher attack success rate with a smaller number of iterations. 
\begin{figure}
	\centering
	\includegraphics[width=\linewidth]{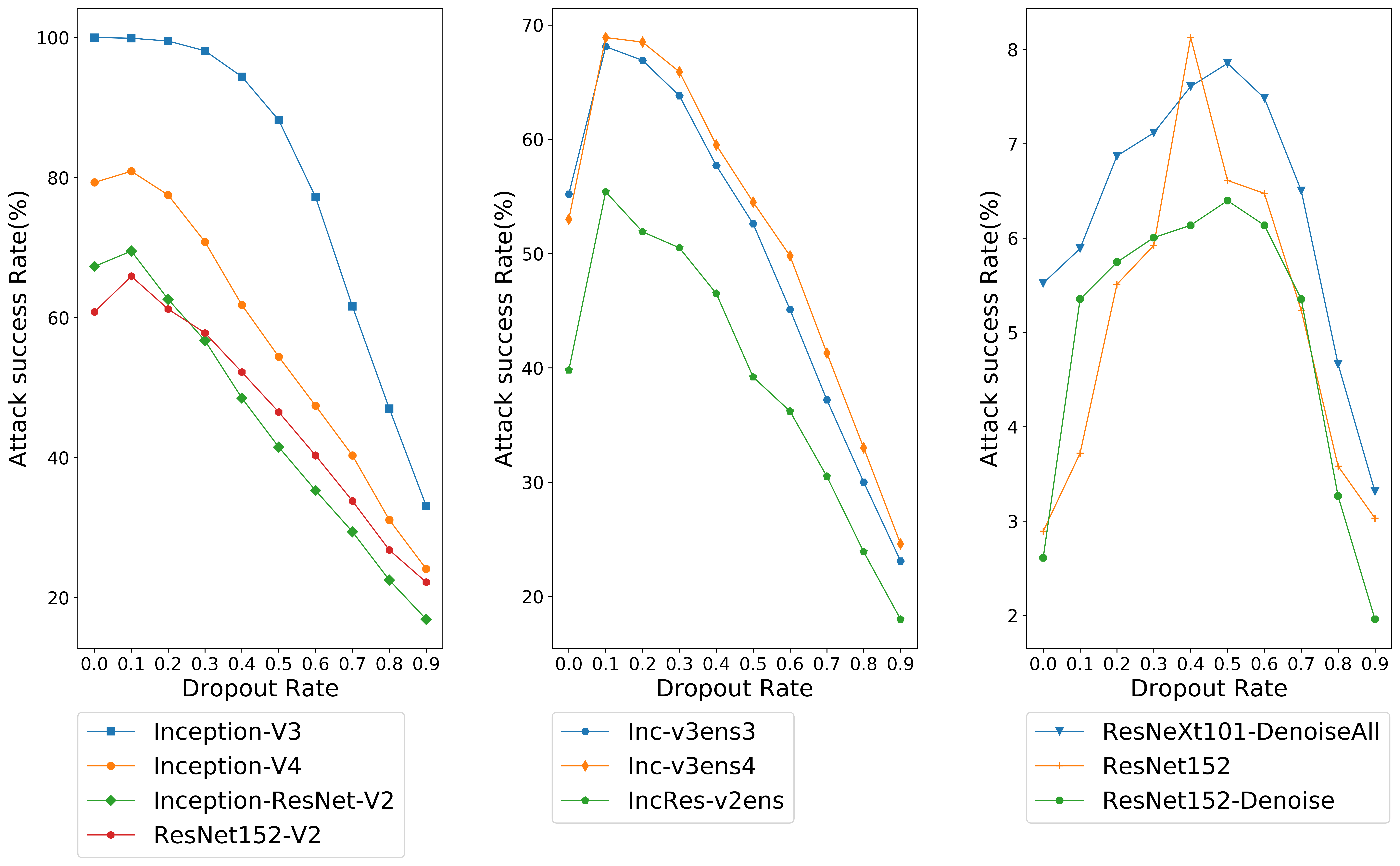}
	\caption{The average attack success rate(\%) of Inc-v3, Inc-v4, IncRes-v2, Res-152, Inc-v${{\rm{3}}_{ens3}}$, Inc-v${{\rm{3}}_{ens4}}$, IncRes-v${2_{{\rm{ens}}}}$, Res15${2_B}$, Res15${2_D}$ and ResNeX${{\rm{t}}_{DA}}$. The adversarial examples are made by DI-TI-ID-MI-FGSM algorithm, and the substitute model is Inc-v3.}
	\label{icml-dropout}
\end{figure}
For non-defense model, the success rate of large-step attack is better than that of small-step attack, but for defense model, the success rate of small-step attack is better than that of large-step attack. We believe that this is because the similarity of the loss surface between the substitute model and the non-defense model is high, so even a large step size can find the extreme point, while the similarity of the loss surface between the substitute model and the defense model is low, and only a small step size can find the appropriate extreme point. However, it should be pointed out that for DI-TI-MI-FGSM, as the number of iterations increases, its performance on the defense model will first increase and then decrease. We believe that this is because the small step size can find a better extreme point at the beginning, but it is easy to fall into over-fitting with the increase of iterations. Our method DI-TI-ID-MI-FGSM can well overcome this problem. The results show that our DI-TI-ID-MI-FGSM improves the attack performance on the defense model by 15.5 \% at 200 times compared with 10 times. \\
Based on the above analysis, we believe that large step size has the advantage of fast convergence, and its attack performance on the non-defense model is more advantageous, but its attack performance on the defense model is weaker than that on the small step size. However, small strides must improve attack performance by extending iteration rounds, which takes a lot of time. In order to maximize the attack performance of the algorithm and prevent excessive time consumption, we set the number of iterations to 50 and the step size to 1.6. Our subsequent experiments will be based on this iterative framework.

\begin{table}[t]
	\centering
	\caption{Success rate of non-target attack (\%) for single model. The leftmost column is an alternative model, and the top row represents the test model ( ' * ' represents a white box attack ).  The adversarial examples are generated by I-FGSM, D${{\rm{I}}^2}$-FGSM, IDI-FGSM, DI-TI-MI-FGSM, DI-TI-ID-MI-FGSM, and DI-TI-IDE-MI-FGSM, respectively.}
	\resizebox{\textwidth}{!}{
				\begin{tabular}{l|l|ccccccc}
					\toprule
					Model & Attack  & Inc-v3 & Inc-v4 & IncRes-v2 &Res-152 &Inc-v${{\rm{3}}_{ens3}}$& Inc-v${{\rm{3}}_{ens4}}$ &IncRes-v${2_{{\rm{ens}}}}$ \\
					\midrule
					\midrule
					\multirow{6}*{Inc-v3}		   
					& I-FGSM & \textbf{100.0*} & 31.2 & 21.1 & 19.3 & 11.2 & 12.7 & 5.6 \\
					& D${{\rm{I}}^2}$-FGSM & \textbf{100.0*} & 53.4 &	43.2 &	36.9 &	15.0  &	14.2 &	7.0 \\
					& IDI-FGSM(Ours) & \textbf{100.0*} & 52.8&	47.3&	37.9&	19.6&	17.5&	10.1  \\
					& DI-TI-MI-FGSM & 98.7* & 67.3& 	55.7& 	51.9& 	50.5& 	50.8& 	39.4 \\
					& DI-TI-ID-MI-FGSM(Ours) & 99.9* & 80.5&	69.3&	68&	69.3&	69.7&	54.3\\
					& DI-TI-IDE-MI-FGSM(Ours) & \textbf{100.0*} & \textbf{89.3} & \textbf{80.0} & \textbf{73.3} & \textbf{72.0} & \textbf{71.6} & \textbf{57.5}\\
					\midrule
					\multirow{6}*{Inc-v4}		   
					& I-FGSM &  44.8  &	\textbf{100.0*}&	25.0 	&25.6	&12.9&	12.2&	6.9 \\
					& D${{\rm{I}}^2}$-FGSM  & 66.4 &	\textbf{100.0*} &	49.1 &	40.9&	15.0 &	15.9&	8.2\\
					& IDI-FGSM(Ours) & 67.9	&\textbf{100.0*}&	52.0 	&44.8&	21.7&	20.5&	13.1  \\
					& DI-TI-MI-FGSM & 73.7&	99.4*&	61.0 &	52.7&	52.6&	51.7&	41.9\\
					& DI-TI-ID-MI-FGSM(Ours) & 86.0 &	\textbf{100.0*}&	72.5&	67.6&	69.1&	67.8&	58.8\\
					& DI-TI-IDE-MI-FGSM(Ours) & \textbf{91.7}&	\textbf{100.0*}&	\textbf{82.3}&	\textbf{76.2}&	\textbf{76.1}&	\textbf{75.1}&	\textbf{63.0}\\	
					\midrule
					\multirow{6}*{IncRes-v2}		   				
					&   I-FGSM &  46.5 &  38.2 &  99.9* &  26.4 &  12.6 &  12.9 &  6.5 \\
					&   D${{\rm{I}}^2}$-FGSM &  73.5 &  68.8 &  99.4* &  51.0  &  19.0  &  19.5 &  12.4\\
					&  IDI-FGSM(Ours) &  74.9 &  65.2 &  \textbf{100.0*} &  51.1 &  22.7 &  21.0  &  15.0 \\
					&   DI-TI-MI-FGSM &  77.7 &  73.5 &  97.0* &  63.5 &  58.4 &  57.6 &  49.3\\
					&   DI-TI-ID-MI-FGSM(Ours) &  88.4 &  \textbf{86.3} &  99.6* &  78.9 &  82.4 &  80.5 &  79.0\\
					&  DI-TI-IDE-MI-FGSM(Ours) &  \textbf{91.7} &  81.6 &  99.9* &  \textbf{85.1} &  \textbf{84.7} &  \textbf{84.3} &  \textbf{81.9}\\    	
					\midrule
					\multirow{6}*{Res-152}	
					&  I-FGSM &  30.3 &  25.8 &  18.7 &  99.5* &  13.8 &  12.4 &  8.2\\
					&  D${{\rm{I}}^2}$-FGSM &  62.1 &  57.5 &  50.5 &  99.2* &  22.7 &  20.6 &  12.0\\
					&  IDI-FGSM(Ours) &  53.8 &  46.4 &  42.7 &  99.6* &  22.4 &  20.2 &  13.7\\
					&  DI-TI-MI-FGSM &  64.0 &  60.0 &  56.4 &  97.9* &  63.8 &  64.1 &  62.6\\
					&   DI-TI-ID-MI-FGSM(Ours) &  75.7 &  72.4 &  68.9 &  99.7* &  77.2 &  76.6 &  69.3\\
					&  DI-TI-IDE-MI-FGSM(Ours) &  \textbf{83.6} &  \textbf{81.7} &  \textbf{79.5} &  \textbf{99.8*} &  \textbf{85.3} &  \textbf{85.3} &  \textbf{80.3}\\	
					\bottomrule
				\end{tabular}}
	\label{sample-table1}%
\end{table}%

\begin{table}[t]
	\centering
	\caption{Success rate of non-target attack (\%) for ensemble model. The leftmost column are attack algorithms, and the top row represents the test model. The adversarial examples are generated by PI-FGSM, DI-TI-MI-FGSM, DI-TI-PI-FGSM, DI-TI-ID-MI-FGSM, DI-TI-ID-PI-FGSM, and DI-TI-IDE-MI-FGSM respectively.}
	\resizebox{\textwidth}{!}{
				\begin{tabular}{l|ccccccc}
					\toprule
					&Inc-v${{\rm{3}}_{ens3}}$& Inc-v${{\rm{3}}_{ens4}}$ &IncRes-v${2_{{\rm{ens}}}}$   & 	ResNeX${{\rm{t}}_{DA}}$	&Res15${2_B}$ &Res15${2_D}$ 
					\\
					\midrule
					\midrule
					PI-FGSM &  67.0 & 67.2 & 59.0 &  8.2 & 6.8 & 7.3 \\
					
					DI-TI-MI-FGSM & 80.0 & 78.8 & 75.6 & 6.3 & 5.7 & 4.3 \\
					
					DI-TI-PI-FGSM & 89.3 & 89.2 & 83.4 & 11.7 & 10.6 & 10.4 \\
					
					DI-TI-ID-MI-FGSM(Ours)  &  96.6 & 96.1 & 93.3 & 10.6 & 10.5 & 10.6\\
					
					DI-TI-ID-PI-FGSM(Ours)  & 88.0 & 89.7 & 84.6 & \textbf{12.6} & \textbf{12.0} & \textbf{12.3}\\
					
					DI-TI-IDE-MI-FGSM(Ours)  & \textbf{96.8} & \textbf{96.8} & \textbf{95.1} & 11.3 & 9.7 & 9.8\\
					\bottomrule
				\end{tabular}}
	\label{sample-table2}%
\end{table}%

\subsection{Dropout rate}\label{sec:3}
In this section, we study the relationship between dropout rate and attack success rate. We use the DI-TI-ID-MI-FGSM algorithm with different dropout, varied from 0.0 to 0.9, to attack Inc-v3 model, with step size as 1.6 and iteration round as 50. The generated adversarial example are then tested in three defenseless models: Inc-v4, IncRes-v2, and Res-152, three ensemble adversarial training models: Inc-v${{\rm{3}}_{ens3}}$, Inc-v${{\rm{3}}_{ens4}}$, and IncRes-v${2_{{\rm{ens}}}}$, and three more robust denoise models: Res15${2_B}$, Res15${2_D}$ and ResNeX${{\rm{t}}_{DA}}$.\\
As shown in the Figure~\ref{icml-dropout}, the results show that dropout rate between 0.0 and 0.2 can improve the transferability of adversarial examples when attacking defenseless model, and the success rate of attack with dropout rate as 0.1 is relatively better. Dropout rate between 0 and 0.5 can improve the attack success rate when attacking ensemble adversarial training models, and dropout rate as 0.1 can achieve a relatively better attack success rate. When attacking denoise models, dropout rate as 0.4 or 0.5 can get better transferability of adversarial examples. 
Increasing dropout rate can prevent the overfitting of adversarial examples to improve the success rate of black-box attack, but too large dropout will lose too much information of adversarial examples, resulting in a reduction in the success rate of white-box attack. Therefore, there is a trade-off on dropout rate. The experimental results show that white-box and black-box have better performance when dropout rate as 0.1. 

\subsection{Single-Model Attacks}\label{sec:4}
In this section, we show the attack performance of I-FGSM, D${{\rm{I}}^2}$-FGSM, IDI-FGSM, DI-TI-MI-FGSM, DI-TI-ID-MI-FGSM, and DI-TI-IDE-MI-FGSM algorithms. Specifically, we chose a single model from Inc-v3, Inc-v4, IncRes-v2, and Res-152 as our substitute model and test it on the remaining three defenseless models and three defense models: Inc-v${{\rm{3}}_{ens3}}$, Inc-v${{\rm{3}}_{ens4}}$, and IncRes-v${2_{{\rm{ens}}}}$. 
As in Table~\ref{sample-table1}, the results show that our method increases by 15.6\% on average compared with I-FGSM, and DI-TI-ID-MI-FGSM increases by 15.4\% on average than DI-TI-MI-FGSM. In addition, our DI-TI-IDE-MI-FGSM achieves an average attack success rate of 79.7\% on the six models, which is 21.4\% higher than DI-TI-MI-FGSM.

\subsection{Ensemble-based Attacks}\label{sec:5}
In this section, we show the attack performance of PI-FGSM, DI-TI-MI-FGSM, DI-TI-PI-FGSM, DI-TI-ID-MI-FGSM, DI-TI-ID-PI-FGSM, and DI-TI-IDE-MI-FGSM algorithms. We attack four normally trained models: Inc-v3, Inc-v4, IncRes-v2, and Res-152 in an ensemble manner to obtain adversarial examples. Then we test them on six defense models: Inc-v${{\rm{3}}_{ens3}}$, Inc-v${{\rm{3}}_{ens4}}$, IncRes-v${2_{{\rm{ens}}}}$, Res15${2_B}$, Res15${2_D}$ and ResNeX${{\rm{t}}_{DA}}$.
As in Table~\ref{sample-table2}, our method combined with DI-TI-PI-FGSM can increase by 1.4\% on average on the denoise model. Our method combined with DI-TI-MI-FGSM can increase by 17.2\% on average on the ensemble adversarial training model. Our method DI-TI-IDE-MI-FGSM can achieve an average attack success rate of 96.2\% on the ensemble adversarial training model, which is higher than the state-of-the-art gradient-based attacks.

\section{Conclusions}
In this paper, we study the relationship between step size, iteration number and maximum disturbance, and explore the potential attack performance of DI-TI-MI-FGSM. We verify that dropout has input invariant, and propose a gradient iterative attack method based on dropout. On this basis, we further propose an integrated version of droput rate. Our method can be combined with other attack methods based on gradient iteration. Our best approach can achieve an average attack success rate of 96.2\% on defense models.

\section{Acknowledgments}
This work was supported by the National Key R${\rm{\& }}$D Program of China under grant 2017YFB1002502 and National Natural Science Foundation of China (No. 61701089, No.61601518 and No. 61372172).


\clearpage

%
%
\bibliographystyle{splncs04}
\bibliography{example_all2}

\begin{thebibliography}{10}
\providecommand{\url}[1]{\texttt{#1}}
\providecommand{\urlprefix}{URL }
\providecommand{\doi}[1]{https://doi.org/#1}

\bibitem{athalye_synthesizing_2018}
Athalye, A., Engstrom, L., Ilyas, A., Kwok, K.: Synthesizing robust adversarial
  examples. In: International conference on machine learning. pp. 284--293.
  PMLR (2018)

\bibitem{baluja_adversarial_2017}
Baluja, S., Fischer, I.: Adversarial transformation networks: {Learning} to
  generate adversarial examples. arXiv preprint arXiv:1703.09387  (2017)

\bibitem{behzadan_vulnerability_2017}
Behzadan, V., Munir, A.: Vulnerability of deep reinforcement learning to policy
  induction attacks. In: International {Conference} on {Machine} {Learning} and
  {Data} {Mining} in {Pattern} {Recognition}. pp. 262--275. Springer (2017)

\bibitem{carlini_towards_2017}
Carlini, N., Wagner, D.: Towards evaluating the robustness of neural networks.
  In: 2017 ieee symposium on security and privacy (sp). pp. 39--57. IEEE (2017)

\bibitem{carlini_audio_2018}
Carlini, N., Wagner, D.: Audio adversarial examples: {Targeted} attacks on
  speech-to-text. In: 2018 {IEEE} {Security} and {Privacy} {Workshops} ({SPW}).
  pp.~1--7. IEEE (2018)

\bibitem{chen_zoo_2017}
Chen, P.Y., Zhang, H., Sharma, Y., Yi, J., Hsieh, C.J.: Zoo: {Zeroth} order
  optimization based black-box attacks to deep neural networks without training
  substitute models. In: Proceedings of the 10th {ACM} workshop on artificial
  intelligence and security. pp. 15--26 (2017)

\bibitem{dong_boosting_2018}
Dong, Y., Liao, F., Pang, T., Su, H., Zhu, J., Hu, X., Li, J.: Boosting
  adversarial attacks with momentum. In: Proceedings of the {IEEE} conference
  on computer vision and pattern recognition. pp. 9185--9193 (2018)

\bibitem{dong_evading_2019}
Dong, Y., Pang, T., Su, H., Zhu, J.: Evading defenses to transferable
  adversarial examples by translation-invariant attacks. In: Proceedings of the
  {IEEE}/{CVF} {Conference} on {Computer} {Vision} and {Pattern} {Recognition}.
  pp. 4312--4321 (2019)

\bibitem{ganin_domain-adversarial_2016}
Ganin, Y., Ustinova, E., Ajakan, H., Germain, P., Larochelle, H., Laviolette,
  F., Marchand, M., Lempitsky, V.: Domain-adversarial training of neural
  networks. The journal of machine learning research  \textbf{17}(1),
  2096--2030 (2016)

\bibitem{gao_patch-wise_2020}
Gao, L., Zhang, Q., Song, J., Liu, X., Shen, H.T.: Patch-wise attack for
  fooling deep neural network. In: European {Conference} on {Computer}
  {Vision}. pp. 307--322. Springer (2020)

\bibitem{ssm}
Gao, L., Zhang, Q., Zhu, X., Song, J., Shen, H.T.: Staircase sign method for
  boosting adversarial attacks. arXiv preprint arXiv:2104.09722  (2021)

\bibitem{goodfellow_explaining_2014}
Goodfellow, I.J., Shlens, J., Szegedy, C.: Explaining and harnessing
  adversarial examples. arXiv preprint arXiv:1412.6572  (2014)

\bibitem{he_deep_2016}
He, K., Zhang, X., Ren, S., Sun, J.: Deep residual learning for image
  recognition. In: Proceedings of the {IEEE} conference on computer vision and
  pattern recognition. pp. 770--778 (2016)

\bibitem{hinton_deep_2012}
Hinton, G., Deng, L., Yu, D., Dahl, G.E., Mohamed, A.r., Jaitly, N., Senior,
  A., Vanhoucke, V., Nguyen, P., Sainath, T.N., {others}: Deep neural networks
  for acoustic modeling in speech recognition: {The} shared views of four
  research groups. IEEE Signal processing magazine  \textbf{29}(6),  82--97
  (2012)

\bibitem{jin_is_2020}
Jin, D., Jin, Z., Zhou, J.T., Szolovits, P.: Is bert really robust? a strong
  baseline for natural language attack on text classification and entailment.
  In: Proceedings of the {AAAI} conference on artificial intelligence. vol.~34,
  pp. 8018--8025 (2020)

\bibitem{komkov_advhat_2021}
Komkov, S., Petiushko, A.: Advhat: {Real}-world adversarial attack on arcface
  face id system. In: 2020 25th {International} {Conference} on {Pattern}
  {Recognition} ({ICPR}). pp. 819--826. IEEE (2021)

\bibitem{kurakin_adversarial_2016}
Kurakin, A., Goodfellow, I., Bengio, S., {others}: Adversarial examples in the
  physical world (2016)

\bibitem{li_learning_2020}
Li, Y., Bai, S., Zhou, Y., Xie, C., Zhang, Z., Yuille, A.: Learning
  transferable adversarial examples via ghost networks. In: Proceedings of the
  {AAAI} {Conference} on {Artificial} {Intelligence}. vol.~34, pp. 11458--11465
  (2020)

\bibitem{lin_nesterov_2019}
Lin, J., Song, C., He, K., Wang, L., Hopcroft, J.E.: Nesterov accelerated
  gradient and scale invariance for adversarial attacks. arXiv preprint
  arXiv:1908.06281  (2019)

\bibitem{moosavi-dezfooli_deepfool_2016}
Moosavi-Dezfooli, S.M., Fawzi, A., Frossard, P.: Deepfool: a simple and
  accurate method to fool deep neural networks. In: Proceedings of the {IEEE}
  conference on computer vision and pattern recognition. pp. 2574--2582 (2016)

\bibitem{nguyen_adversarial_2020}
Nguyen, D.L., Arora, S.S., Wu, Y., Yang, H.: Adversarial {Light} {Projection}
  {Attacks} on {Face} {Recognition} {Systems}: {A} {Feasibility} {Study}. In:
  Proceedings of the {IEEE}/{CVF} {Conference} on {Computer} {Vision} and
  {Pattern} {Recognition} {Workshops}. pp. 814--815 (2020)

\bibitem{redmon_yolov3_2018}
Redmon, J., Farhadi, A.: Yolov3: {An} incremental improvement. arXiv preprint
  arXiv:1804.02767  (2018)

\bibitem{srivastava_dropout_2014}
Srivastava, N., Hinton, G., Krizhevsky, A., Sutskever, I., Salakhutdinov, R.:
  Dropout: a simple way to prevent neural networks from overfitting. The
  journal of machine learning research  \textbf{15}(1),  1929--1958 (2014)

\bibitem{sutskever_sequence_2014}
Sutskever, I., Vinyals, O., Le, Q.V.: Sequence to sequence learning with neural
  networks. arXiv preprint arXiv:1409.3215  (2014)

\bibitem{szegedy_inception-v4_2017}
Szegedy, C., Ioffe, S., Vanhoucke, V., Alemi, A.: Inception-v4,
  inception-resnet and the impact of residual connections on learning. In:
  Proceedings of the {AAAI} {Conference} on {Artificial} {Intelligence}.
  vol.~31 (2017)

\bibitem{szegedy_rethinking_2016}
Szegedy, C., Vanhoucke, V., Ioffe, S., Shlens, J., Wojna, Z.: Rethinking the
  inception architecture for computer vision. In: Proceedings of the {IEEE}
  conference on computer vision and pattern recognition. pp. 2818--2826 (2016)

\bibitem{szegedy_intriguing_2013}
Szegedy, C., Zaremba, W., Sutskever, I., Bruna, J., Erhan, D., Goodfellow, I.,
  Fergus, R.: Intriguing properties of neural networks. arXiv preprint
  arXiv:1312.6199  (2013)

\bibitem{tramer_ensemble_2017}
Tramèr, F., Kurakin, A., Papernot, N., Goodfellow, I., Boneh, D., McDaniel,
  P.: Ensemble adversarial training: {Attacks} and defenses. arXiv preprint
  arXiv:1705.07204  (2017)

\bibitem{wang_defensive_2018}
Wang, S., Wang, X., Zhao, P., Wen, W., Kaeli, D., Chin, P., Lin, X.: Defensive
  dropout for hardening deep neural networks under adversarial attacks. In:
  Proceedings of the {International} {Conference} on {Computer}-{Aided}
  {Design}. pp.~1--8 (2018)

\bibitem{wu_understanding_2018}
Wu, L., Zhu, Z., Tai, C., {others}: Understanding and enhancing the
  transferability of adversarial examples. arXiv preprint arXiv:1802.09707
  (2018)

\bibitem{xiao_generating_2018}
Xiao, C., Li, B., Zhu, J.Y., He, W., Liu, M., Song, D.: Generating adversarial
  examples with adversarial networks. arXiv preprint arXiv:1801.02610  (2018)

\bibitem{xie_feature_2019}
Xie, C., Wu, Y., Maaten, L.v.d., Yuille, A.L., He, K.: Feature denoising for
  improving adversarial robustness. In: Proceedings of the {IEEE}/{CVF}
  {Conference} on {Computer} {Vision} and {Pattern} {Recognition}. pp. 501--509
  (2019)

\bibitem{xie_improving_2019}
Xie, C., Zhang, Z., Zhou, Y., Bai, S., Wang, J., Ren, Z., Yuille, A.L.:
  Improving transferability of adversarial examples with input diversity. In:
  Proceedings of the {IEEE}/{CVF} {Conference} on {Computer} {Vision} and
  {Pattern} {Recognition}. pp. 2730--2739 (2019)

\bibitem{xu_adversarial_2020}
Xu, K., Zhang, G., Liu, S., Fan, Q., Sun, M., Chen, H., Chen, P.Y., Wang, Y.,
  Lin, X.: Adversarial t-shirt! evading person detectors in a physical world.
  In: European {Conference} on {Computer} {Vision}. pp. 665--681. Springer
  (2020)

\bibitem{zou_improving_2020}
Zou, J., Pan, Z., Qiu, J., Liu, X., Rui, T., Li, W.: Improving the
  {Transferability} of {Adversarial} {Examples} with
  {Resized}-{Diverse}-{Inputs}, {Diversity}-{Ensemble} and {Region} {Fitting}.
  In: European {Conference} on {Computer} {Vision}. pp. 563--579. Springer
  (2020)

\end{thebibliography}
\clearpage

\end{document}